\documentclass[letterpaper, 10 pt, conference]{ieeeconf}

\IEEEoverridecommandlockouts                              
\usepackage{epsfig}
\usepackage{epstopdf}
\usepackage{graphics}
\usepackage{upgreek}
\usepackage{amssymb}
\usepackage{amsfonts}
\usepackage{amsmath}
\usepackage{refstyle}
\usepackage{xcolor}
\usepackage{subfigure}
\usepackage{todonotes}
\usepackage{mathtools}
\usepackage{algorithm}
\usepackage{algpseudocode}
\usepackage{cite}
\usepackage{balance}
\usepackage{mathrsfs}
\usepackage{lipsum}
\usepackage{multicol}

\title{\LARGE \bf
ColibriDoc:  An Eye-in-Hand  Autonomous Trocar Docking System 
}

\author{Shervin Dehghani$^{1*}$, Michael Sommersperger$^{1*}$, Junjie Yang$^{2}$, Benjamin Busam$^{1}$, Kai Huang$^{3}$ \\
Peter Gehlbach$^{4}$, Iulian Iordachita$^{5}$, Nassir Navab$^{1,5}$ and M. Ali Nasseri$^{1,2}$
\thanks{* The first two authors contributed equally to this paper. Corresponding author: M. Ali Nasseri \tt\small (ali.nasseri@mri.tum.de)}
\thanks{$^{1}$ S. Dehghani, M. Sommersperger, B. Busam, N. Navab and M. Ali Nasseri are with Department of Computer Science in Technische Universit\"{a}t M\"{u}nchen, M\"{u}nchen 85748 Germany. 
}%
\thanks{$^{2}$ J. Yang and M. Ali Nasseri are with Augenklinik und Poliklinik, Klinikum rechts der Isar derc Technische Universit\"{a}t M\"{u}nchen, M\"{u}nchen 81675 Germany.
        }%
\thanks{$^{3}$K. Huang is with Key Laboratory of Machine Intelligence and Advanced Computing (Sun Yat-sen University), Guangzhou, China. 
        }%
\thanks{$^{4}$P. Gehlbach is with Wilmer Eye Institute, Johns Hopkins Hospital, Baltimore, MD, USA.
        }%
\thanks{$^{5}$I. Iordachita and N. Navab are with Laboratory for Computational Sensing and Robotics, Johns Hopkins University, Baltimore, MD, USA.
        }%
}
\DeclarePairedDelimiterX{\norm}[1]{\lVert}{\rVert}{#1}
\begin{document}
\definecolor{ben}{rgb}{0.9,0.,0.5}
\newcommand{\ben}[1]{\textcolor{ben}{\emph{Ben:~{#1}}}}
\definecolor{sher}{rgb}{0.0,0.3,0.5}
\newcommand{\sher}[1]{\textcolor{sher}{\emph{Sher:~{#1}}}}
\definecolor{michi}{rgb}{0.3,0.1,0.5}
\newcommand{\michi}[1]{\textcolor{michi}{\emph{Michi:~{#1}}}}
\definecolor{ali}{rgb}{0.3,0.8,0.5}
\newcommand{\ali}[1]{\textcolor{ali}{\emph{Ali:~{#1}}}}

\newcommand*{\vertbar}{\rule[-1ex]{0.5pt}{2.5ex}}
\newcommand*{\horzbar}{\rule[.5ex]{2.5ex}{0.5pt}}

\maketitle


\begin{abstract}

Retinal surgery is a complex medical procedure that requires exceptional expertise and dexterity. For this purpose, several robotic platforms are currently being developed to enable or improve the outcome of microsurgical tasks.
Since the control of such robots is often designed for navigation inside the eye in proximity to the retina, successful trocar docking and inserting the instrument into the eye represents an additional cognitive effort, and is therefore one of the open challenges in robotic retinal surgery.
For this purpose, we present a platform for autonomous trocar docking that combines computer vision and a robotic setup.
Inspired by the Cuban Colibri (hummingbird) aligning its beak to a flower using only vision, we mount a camera onto the endeffector of a robotic system.
By estimating the position and pose of the trocar, the robot is able to autonomously align and navigate the instrument towards the Trocar's Entry Point (TEP) and finally perform the insertion. 
Our experiments show that the proposed method is able to accurately estimate the position and pose of the trocar and achieve repeatable autonomous docking. The aim of this work is to reduce the complexity of robotic setup preparation prior to the surgical task and therefore, increase the intuitiveness of the system integration into the clinical workflow.  

\end{abstract}

\bstctlcite{IEEEexample:BSTcontrol}

\begin{keywords}

Medical Robots and Systems; Surgical Robotics: Planning; Computer Vision for Medical Robotics.  
\end{keywords}

\markboth{IEEE Robotics and Automation Letters. Preprint Version. Accepted XXXX, 202X}
{Dehghani \MakeLowercase{\textit{et al.}}: AutroDoc: Autonomous Trocar Docking for Vitreoretinal Surgery using a Robot-Mounted Camera}
%
%
%
%
\section{Introduction}
\begin{figure}[H]
    \centering
    \includegraphics[width=0.99\linewidth]{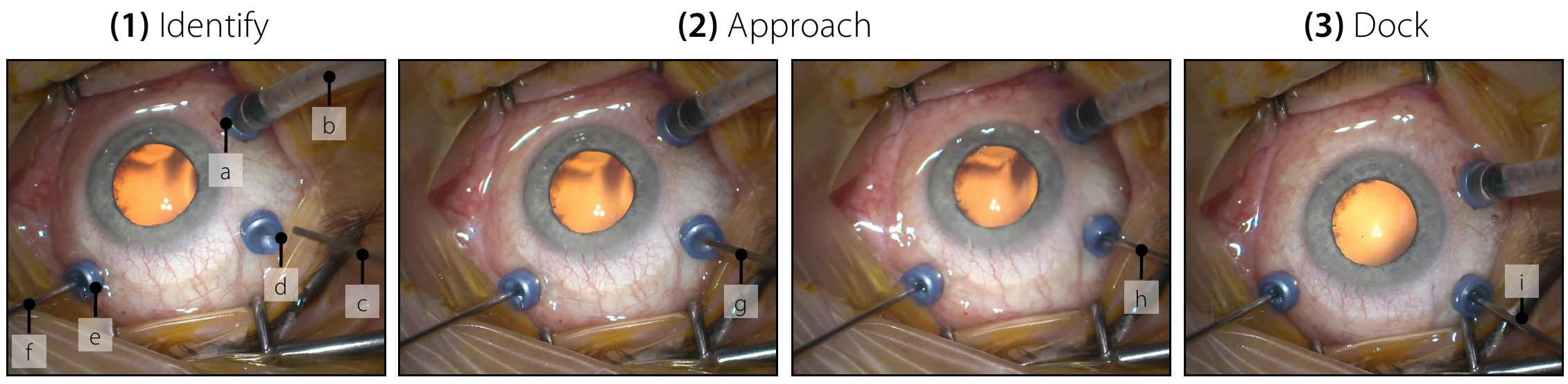}
\caption{
In conventional vitreoretinal interventions, trocars are inserted into the sclera to dock various surgical instruments and allow access to the surgical site: \textbf{a.} Trocar I \textbf{b.} Infusion line \textbf{c.} Instrument approaching the trocar \textbf{d.} Trocar II \textbf{e.} Trocar III \textbf{f.} Illuminator \textbf{g.} Instrument docking the trocar \textbf{h.} Instrument aligning towards the trocar orientation \textbf{i.} Properly aligned instrument inserted into the trocar.}
\label{fig:intro}
\vspace{-0.4cm}
\end{figure}

Vitreoretinal surgery is known to be one of the most challenging and delicate surgical procedures, requiring surgeons to have sufficient expertise and exceptional hand stability to manipulate microsurgical instruments. 
The demand for specialized retinal surgeons is high, as more than 300 million~\cite{de2020cutting} patients are affected by visual disorders caused by various retinal diseases.
These diseases are mainly treated through vitreoretinal interventions aiming to preserve or restore vision. 
Since vitreoretinal surgery is a minimally invasive procedure, trocars are placed as insertion ports on the sclera (sclerotomy) prior to surgery allowing access to the operating area. 
The surgeon then docks an infusion line, an illuminator and a surgical instrument into these ports to start the vitreoretinal intervention (see Fig.~\ref{fig:intro}).
In conventional vitreoretinal interventions, surgeons rely on both their visual and haptic feedback to introduce instruments through the trocars. 
They approach by visually identifying the trocar and fine-tune the insertion alignment by sensing the forces during the docking procedure.

In recent years, robotic platforms have made promising strides towards facilitating and improving vitreoretinal procedures and could enable more surgeons to perform such complicated tasks~\cite{yang2017medical,lu2018vision,vander2020robotic}. 
Despite their significant technical improvements, robotic systems are still limited in interfacing and integration capabilities into the surgical workflow.
For instance, in current robotic setups clinicians need to invest significant time and energy for the preparation of the system and the accurate manual positioning of the robot in order to adjust the appropriate docking orientation prior to the main procedure. 
Therefore, in minimally invasive robotic surgery, and specifically in vitreoretinal robotic surgery, the task of robot positioning, trocar docking, and instrument insertion poses additional cognitive demands on the surgeon.
The main reason for this effort is that the control of such systems is designed for delicate procedures at micron-level scale inside the eye, with limited working volume, and is often manipulated using a controller such as a joystick~\cite{Zhou2020}.
Therefore, inserting the microsurgical instrument into the trocar using this control system naturally becomes more challenging and time-consuming compared to the conventional insertion in manual surgery.
Automating the task of navigating a surgical tool from a safe distance towards the trocar and introducing it into the eye could thus relieve additional cognitive load on the surgeon and consequently reduce the complexity of employing a robotic system in the clinical workflow.

In this paper, we propose \textit{ColibriDoc}, a system for autonomous trocar docking and instrument insertion. Our setup consists of an RGB camera, mounted on the endeffector of an ophthalmic robot. Its design is mainly inspired by the natural behavior of Cuban Colibris (hummingbirds), which hover at flowers and rely only on their stereo vision to align and insert their beak~\cite{goller2014hummingbirds, goller2017visual}.
Due to the space constraints in the operating room and the limited mounting options on robotic systems, contrary to the hummingbird we follow a mono-vision approach based on a single camera.
\begin{figure}[t]
\vspace{0.2cm}
    \centering
    \includegraphics[width=1\linewidth]{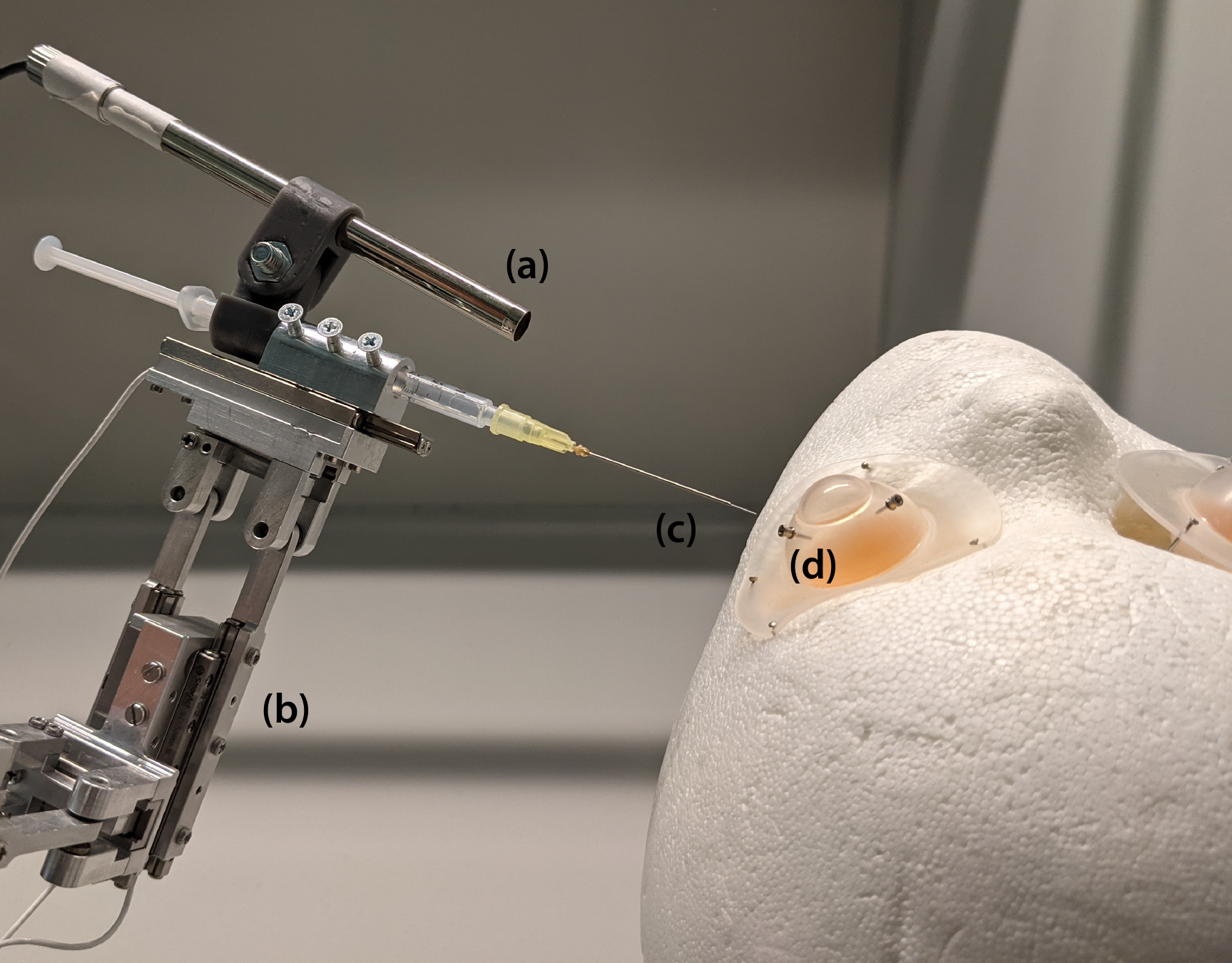}
    \caption{Our setup for autonomous trocar docking consists of a mono-vision RGB camera (a) mounted on a microsurgical robot (b). The surgical instrument (c) is attached to the robot endeffector and autonomously navigated to perform the docking procedure to the target trocar (d).}
    \label{fig:setup}
\vspace{-0.4cm}
\end{figure}

Our proposed method precisely identifies the homogeneous location of the Trocar Entry Point (TEP) and its orientation from RGB images using a two-stage learning-based approach.
The robot then autonomously aligns the surgical tool with the trocar orientation and ultimately performs trocar docking and instrument insertion.
For a proof of concept and experimental evaluation, we implement the proposed method on a robotic system designed for vitreoretinal surgery, which is described in~\cite{Zhou2020}.
Our experiments show that the TEP and trocar pose can accurately be determined using RGB images only.
Most importantly, we show that such a system is capable of performing trocar docking and instrument insertion in a precise and repeatable manner, automating a task that in robotic surgery poses additional complexity on the surgeon.
We demonstrate the concept for vitreoretinal surgery, however, this methodology could also be transferred to other types of minimally invasive robotic surgery.
To the best of our knowledge, the system proposed in this paper is the first work towards automating the trocar docking procedure. 
\section{Related works}
\label{sec:Relatedwork}
\textbf{Robotic Systems:} 
In the last decade, many robotic systems have been designed and applied in surgeries as robotic assistants to promote autonomy and high accuracy such as~\cite{ullrich2013mobility,rahimy2013robot,gijbels2014experimental}. These approaches are classified into the following types according to the interaction between the surgeon and the robotic system: 1) hand-held surgery instruments~\cite{song2012active} fully controlled by the surgeon, 2) cooperatively controlled robotic systems~\cite{Taylor1999} in which the surgical instrument is jointly controlled by the robot and surgeon and 3) teleoperation systems, in which the robot is remotely controlled by the surgeon via a guidance device such as a joystick~\cite{zhou2019towards,molaei2017toward}. 
In this work we employ a 5 DoF hybrid parallel-serial robot~\cite{zhou2019towards, nasseri2013introduction} specifically designed for delicate vitreoretinal procedures, which can be controlled as a teleoperation system or by a software framework. 

\textbf{Trocar Entry Point Detection and Pose Estimation:} 
Till date, only few works have been published with a motivation similar to the one for autonomous trocar docking proposed in this paper.
Multiple works have focused on estimating the TEP and positioning the Remote Center of Motion (RCM) during surgery using a geometric approach~\cite{Smits2019,Gruijthuijsen2018,Dong2016} or an external stereo-vision system~\cite{Rosa2015}.
Birch et al.~\cite{Birch2021} recently published an initial paper on the development of an instrument with two integrated miniature cameras to detect the trocar position and an internal measurement unit to estimate the RCM point of the robot.
Rather than performing the entire trocar docking procedure at an initial safe distance from the target, such approaches focus on repositioning the instrument to align the RCM with the TEP after instrument insertion.
In contrary, we detect the TEP and its pose at a safe initial distance from the eye and autonomously navigate the robot to perform docking and instrument insertion.

In recent years, 6D Pose Estimation of objects has become a popular research topic. 
While depth-based methods show higher accuracy than monocular-based methods, the recent progress in the field of monocular pose estimation \cite{kehl2017ssd,wang2021gdr,labbe2020cosypose},  demonstrates the high capabilities of using monocular camera instead of stereo or depth cameras.
These approaches can be categorized into two main groups: indirect and direct methods. 
The indirect methods goal is to find $2D-3D$ point correspondences to directly derive the pose from a P$n$P algorithm~\cite{tekin2018real}.
Direct methods, on the other hand, approach the problem as a regression~\cite{labbe2020cosypose, wang2021gdr} or classification~\cite{kehl2017ssd} task using a direct differentiable method.
To estimate the pose of a microsurgical trocar as part of our proposed system, we use a mono-camera based direct approach, in order to better cope with the texture-less and symmetric properties of trocar and the limitations in the workspace of the robot, as described in more detail in section~\ref{sec:4DoFTrocarPoseEstimation}.

\section{Dataset}
\label{sec:dataset}
\begin{figure}[H]
    \centering
    \includegraphics[width=0.9\linewidth]{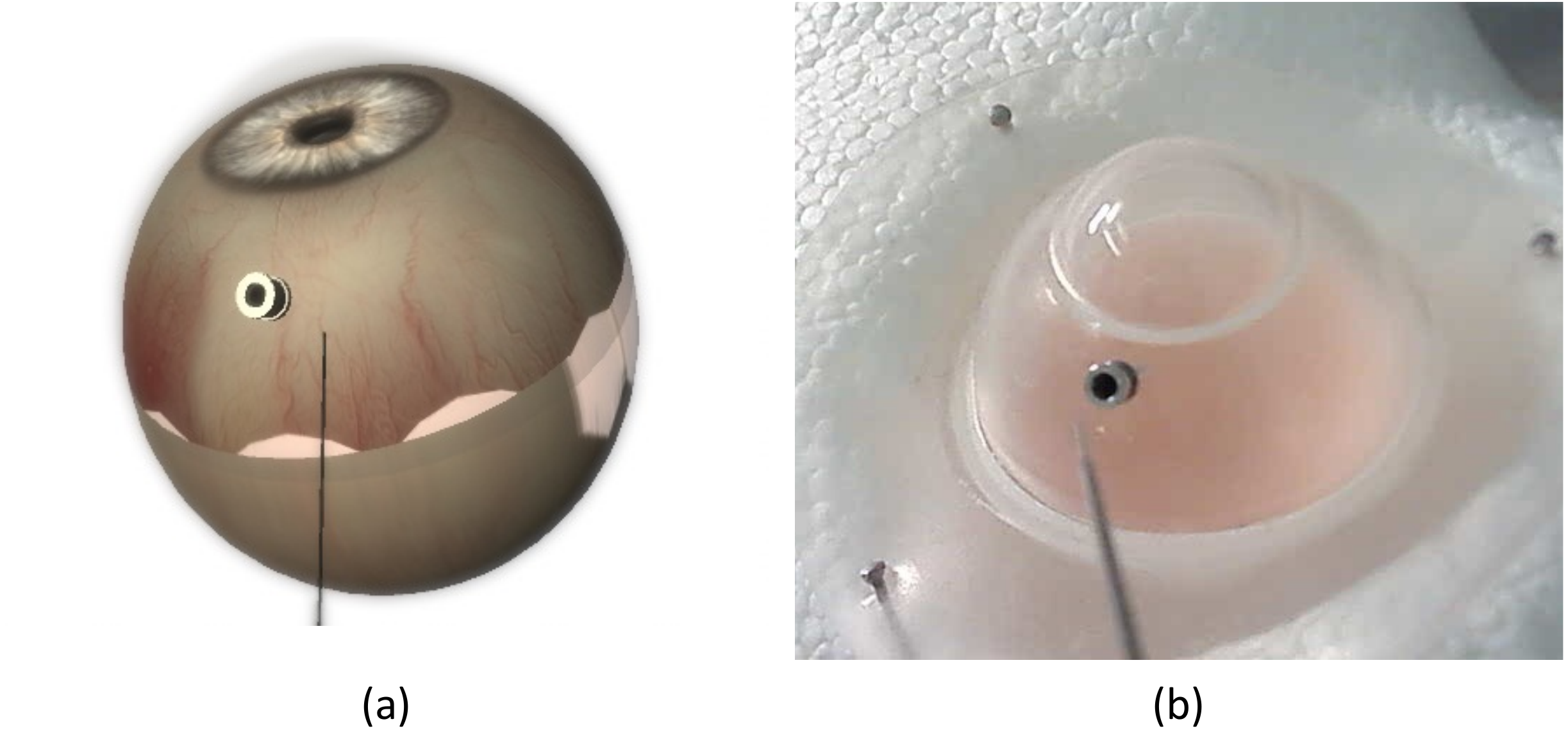}
    \caption{A synthetic (a) and a real (b) image with similar trocar poses.}
    \label{fig:DataSetSynthReal}
\vspace{-0.4cm}
\end{figure}

A major challenge in the learning-based detection of the trocar and the estimation of its pose is the acquisition of a dataset with accurate ground truth information containing both the location and orientation of the TEP.
Due to the lack of available datasets, we generate a purely synthetic dataset from a virtual setup, which contains virtual models of an eye, a trocar, a surgical needle, as well as a virtual camera.
The camera and the tool are positioned relative to the eye and the trocar similar to the real setup. 
In this virtual environment the exact location and orientation of the trocar in relation to the camera is known.
The parameters of the virtual camera are adjusted according to the parameters of the calibrated camera that is mounted on the real robotic system and the resolution of the rendered frames is adapted to match the image resolution of the robot-mounted camera.
To generate the synthetic dataset, we randomize the position and orientation of the trocar in the rendered frames, along with the location of the trocar on the eye model.
Furthermore, we vary the metallic properties and glossiness of the trocar material, as well as the direction and intensity of the lighting in the scene.
In 20\% of the acquired frames the eye model is not rendered to avoid bias towards specific eye characteristics, but rather focus on learning the features of the trocar.
In an effort to improve robustness towards changing peripheral areas when training deep learning models to estimate the TEP and pose, randomized images from the cocodataset~\cite{lin2014microsoft} are rendered in the background of the virtual scene.
The generated synthetic dataset $DS_{synthetic}$ contains 2000 images with ground truth information of the TEP in image coordinates and the 3D pose of the trocar in the virtual scene relative to the camera.
For fine-tuning and bridging the gap between the synthetic and real data domain, we acquire two different labeled datasets, consisting of images extracted from videos captured with our proposed system.
One of them, $DS_{trainTEP}$, is used for the TEP detection, while the other one, $DS_{trainPoseReal}$, is used for orientation estimation. To acquire these datasets, a microsurgical trocar (23G from RETILOCK) was inserted into a phantom eye and the robotic setup was positioned in a realistic distance from the trocar. Two different phantom eyes (VR Eye from Philips Studio and BIONIKO eye model with flex orbit holder) were used to replicate real surgical scenarios.
Fig.~\ref{fig:DataSetSynthReal} shows an example of a synthetic and a real image.
For $DS_{trainTEP}$, six videos were acquired, from which 1280 frames were extracted to create the dataset. The ground truth location of the TEP was manually annotated by a biomedical engineering expert. 
$DS_{trainPoseReal}$ consists of 100 images, in which we obtain the ground truth orientation of the trocar in relation to the camera using a marker aligned with the trocar.
With the same approaches, 160 images were extracted creating the dataset $DS_{testTEP}$ and 50 images creating $DS_{testPoseReal}$, which are used for the evaluations presented in section \ref{sec:experiment}.

\section{Method}

\label{sec:method}
In this section, the proposed method for autonomous trocar docking is described in detail. 
Firstly, the setup consisting of an ophthalmic robot and a mono-vision camera mounted on the robot endeffector is illustrated.
Thereafter, the components to detect the position and orientation of the trocar, as well as the post-processing steps to refine the estimations are presented.
Finally, we outline the trajectory planning to autonomously align the instrument with the trocar orientation and navigate the tooltip towards the TEP to perform the docking procedure.
\begin{figure}[t]
    \vspace{0.2cm}
    \centering
    \includegraphics[width=1\linewidth]{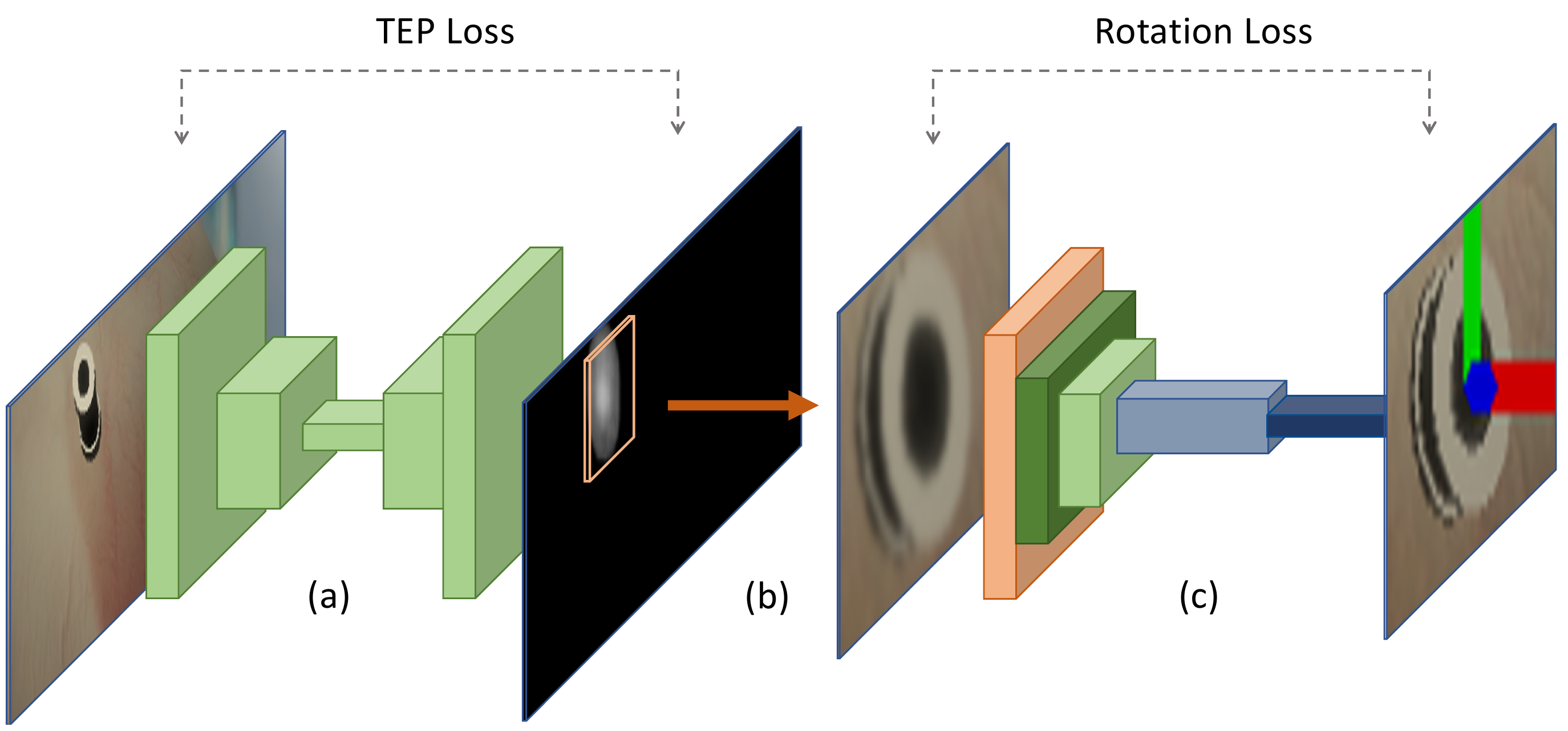}
    \caption{The proposed pipeline for the 4DoF trocar pose estimation. \textbf{a.} A U-Net like network to segment the TEP, \textbf{b.} Cropping the image around TEP, \textbf{c.} A Resnet34 to regress the pose.}
    \label{fig:pipeline}
\vspace{-0.4cm}
\end{figure}

\subsection{Setup}
In this work, a 5DoF robotic micromanipulator~\cite{nasseri2013introduction, zhou2018towards} is employed.
The robot consists of two parallel coupled joint mechanisms for translation and rotation in two axes and a decoupled prismatic joint for $Z$ movement of the endeffector.
All joints are actuated by micron-precision piezo motors with integrated sub-micron optical encoders. 
To achieve autonomous trocar docking, a Teslong Multi-Function Soldering Magnifier Camera is calibrated and rigidly mounted on the endeffector of the robot.
A sub-retinal cannula (23G with 40G tip) is attached to the robot endeffector and a 3D printed holder was specifically designed to mount the camera to the syringe in a suitable orientation.
The proposed robotic setup is illustrated in Fig.~\ref{fig:setup}.
For this Eye-in-Hand~\cite{Flandin2000} setup, in which the endeffector moves the instrument along the camera, a Hand-Eye calibration~\cite{horaud1995hand} is performed to acquire the camera pose w.r.t. to the robot.

To create realistic docking targets, conventional 23G trocars are placed 3.5 mm posterior to the limbus of two surgical training phantom eyes (described in section~\ref{sec:dataset}) with proper scleral textures and deformability.


\subsection{4DoF Trocar Pose Estimation}
\label{sec:4DoFTrocarPoseEstimation}
In a typical vitreoretinal surgery, several trocars are placed to provide the ports for instruments and access to the surgical site.
The first step towards autonomous robotic trocar docking is identifying the appropriate entry point and pose of the target trocar based on image frames captured by the camera mounted on the robot.
Our goal is to estimate the projected 2D point of the closest trocar's TEP in the current image plane and the normal vector of its cross-section.
Due to the nature of vitreoretinal surgery and the small working area of the surgical robot, the robot's endeffector is initially positioned in a reachable distance to the trocar.
Thus, the closest trocar is the docking target, which is also directly visible in the camera's field of view.
We employ a two stage 4DoF pose estimation neural network, to address both the TEP and trocar pose estimation from single RGB images. We initially locate the TEP by the first network, which extracts the ROI that is subsequently forwarded to the second network, responsible for estimating the orientation. Obtaining the orientation along the TEP provides the homogeneous location of the trocar, which is the basis of our hummingbird-inspired docking approach, as explained further in section \ref{sec:TrajectoryPlanning}.


\textbf{Trocar Entry Point Detection.} 
Detecting the TEP is, firstly, relevant for extraction of a ROI to be later used for pose estimation, and  controlling the robot to align the tooltip with the trocar orientation and to perform the trocar docking procedure.
In addition, a small region of interest around the TEP is subsequently extracted from the input frame and used to estimate the trocar orientation. Rather than using an object detection network to estimate a bounding box around the trocar, it is instead important to estimate the exact location of the TEP, since, depending on the orientation of the trocar, the entry point cannot easily be obtained from a bounding box.
For this stage, a U-Net-style network \cite{Ronneberger2015} using a Resnet34 \cite{he2015deep} as a backbone feature extractor is trained to obtain the location of the TEP in the camera images.
To generate the ground truth images for training the network, a Gaussian function is applied around the ground truth TEP with a maximum distance of 15 pixels from its center and a sigma of 1.

The network is first pretrained on $DS_{synthetic}$ and afterwards fine-tuned on $DS_{trainTEP}$.
The last network layer uses the sigmoid activation function, which is followed by the binary cross entropy loss.

To derive the final image coordinates of the TEP location for each processed frame, we consider all pixel locations $(x,y)$ in the network output, which satisfy the equation 
\begin{equation}
    pred(x,y) \geq 0.8 * max(pred)
\end{equation}
where $pred(x,y)$ defines the confidence that the output image at the pixel location $(x,y)$ is classified as TEP and $max(pred)$ is the overall maximum value in the network's output image.
The TEP is then estimated as the median of the extracted candidate locations.

During inference, we apply further post-processing which combines the predicted locations of every seven consecutive frames to improve the robustness of the prediction.
First, the median value of all seven predictions is calculated and the Euclidean distance between each individual prediction and the overall median value is determined.
All predictions more than one quarter of a standard deviation distant from the overall median location are considered outliers.
The remaining estimates are then averaged to produce the final TEP.

\textbf{Orientation Estimation.}
While dealing with a texture-less and symmetric object as a trocar, we leverage a direct pose estimation method with symmetric loss to predict the normal vector of trocars' cross-section. 

In order to have a continuous space of rotations in $\mathbf{SO(3)}$, we follow the method introduced in~\cite{Zhou_2019_CVPR} to parameterize rotation angles. As demonstrated in~\cite{labbe2020cosypose,wang2021gdr}, the mapping function $f$ to the 6-dimensional representation $\mathbf{R}_{\text{6d}}$ is defined as the first two columns of $\mathbf{R}$
{\small
\begin{equation}
\quad f\left(\left[ {\begin{array}{ccc}
   \vertbar & \vertbar & \vertbar \\
   R_1 & R_2 & R_3 \\
   \vertbar & \vertbar & \vertbar
  \end{array} } \right]\right) = 
  \left[ {\begin{array}{ccc}
   \vertbar & \vertbar \\
   R_1 & R_2 \\
   \vertbar & \vertbar
  \end{array} } \right]
\label{eqn:gramschmidt_repr}
\end{equation}
}
Given a 6-dimensional vector $\mathbf{R}_{\text{6d}} = [\mathbf{r}_1 | \mathbf{r}_2]$,
the unit and orthogonal rotation matrix $\mathbf{R}$
is computed as
\begin{equation}
\begin{cases}
\mathbf{R}_1 = \phi(\mathbf{r}_1) \\
\mathbf{R}_3 = \phi(\mathbf{R}_1 \times \mathbf{r}_2) \\
\mathbf{R}_2 = \mathbf{R}_3 \times \mathbf{R}_1 \\
\end{cases},
\label{eq:r6_to_rot}
\end{equation}
where $\phi(\bullet)$ denotes the vector normalization operation.

For the convenience of our problem, we use a permutation of columns of $\mathbf{R}$ which changes the 6D representation to
\begin{equation}
\label{eq:r7}
    \mathbf{R}_{\text{6d}} = \left[\mathbf{R}_Z ~|~ \mathbf{R}_Y\right]
\end{equation}

\begin{figure}[H]
    \centering
    \includegraphics[width=0.3\linewidth]{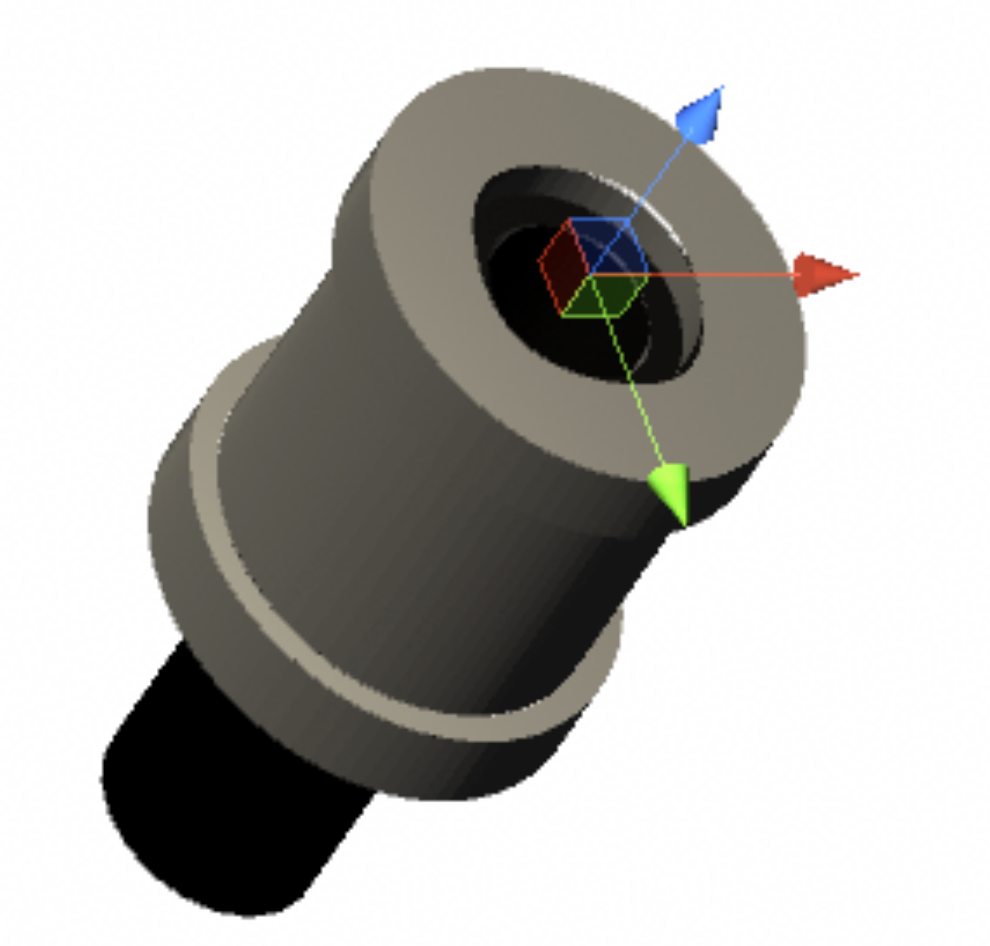}
    \caption{Illustration of the trocar coordinate system. The trocar orientation can be defined only by the rotation around its x and y axis, which are visualized in red and green, respectively.}
    \label{fig:TrocarCoordSystem}
    \vspace{-0.4cm}
\end{figure}

Thereafter, we use a Resnet34-based~\cite{he2015deep} backbone to convert the input image, derived from a trocar-centered ROI extraction into its features, followed by fully connected layers to regress to the 6D representation of the rotation. 
Assuming the coordinate system of a trocar as illustrated in Fig.~\ref{fig:TrocarCoordSystem}, it can be seen that a trocar is symmetric along its $Z$ axis. In this regard, we design a loss which does not penalize the network for non-relevant rotations around the trocar's z axis. The angle between the ground truth $\mathbf{R}_Z$ and estimation can be computed as:

\begin{equation}
\label{eq:deltatheta}
\Delta_{\theta} = \arccos(
{\mathbf{R}}^{gt}_Z
\boldsymbol{\cdot}
{\mathbf{R}}^{pred}_Z)
\end{equation}
For $\ell^2$-normalized vectors $\mathrm{MSE}$ is proportional to $cosine$ distance, and due to this fact we use Eq.~\ref{eq:new_pose_loss} as the loss function, which shows a more promising convergence.
\begin{equation}
\label{eq:new_pose_loss}
\mathcal{L}_{rotation} = \mathrm{MSE}({\mathbf{R}}^{gt}_Z, {\mathbf{R}}^{pred}_Z)
\end{equation}

In order to be robust against noise, we average seven consecutive estimations using the method introduced in \cite{markley2007averaging}. 

\subsection{Trajectory Planning}
\label{sec:TrajectoryPlanning}
\begin{figure}[t]
    \centering
    \vspace{0.6cm}
    \includegraphics[width=0.9\linewidth]{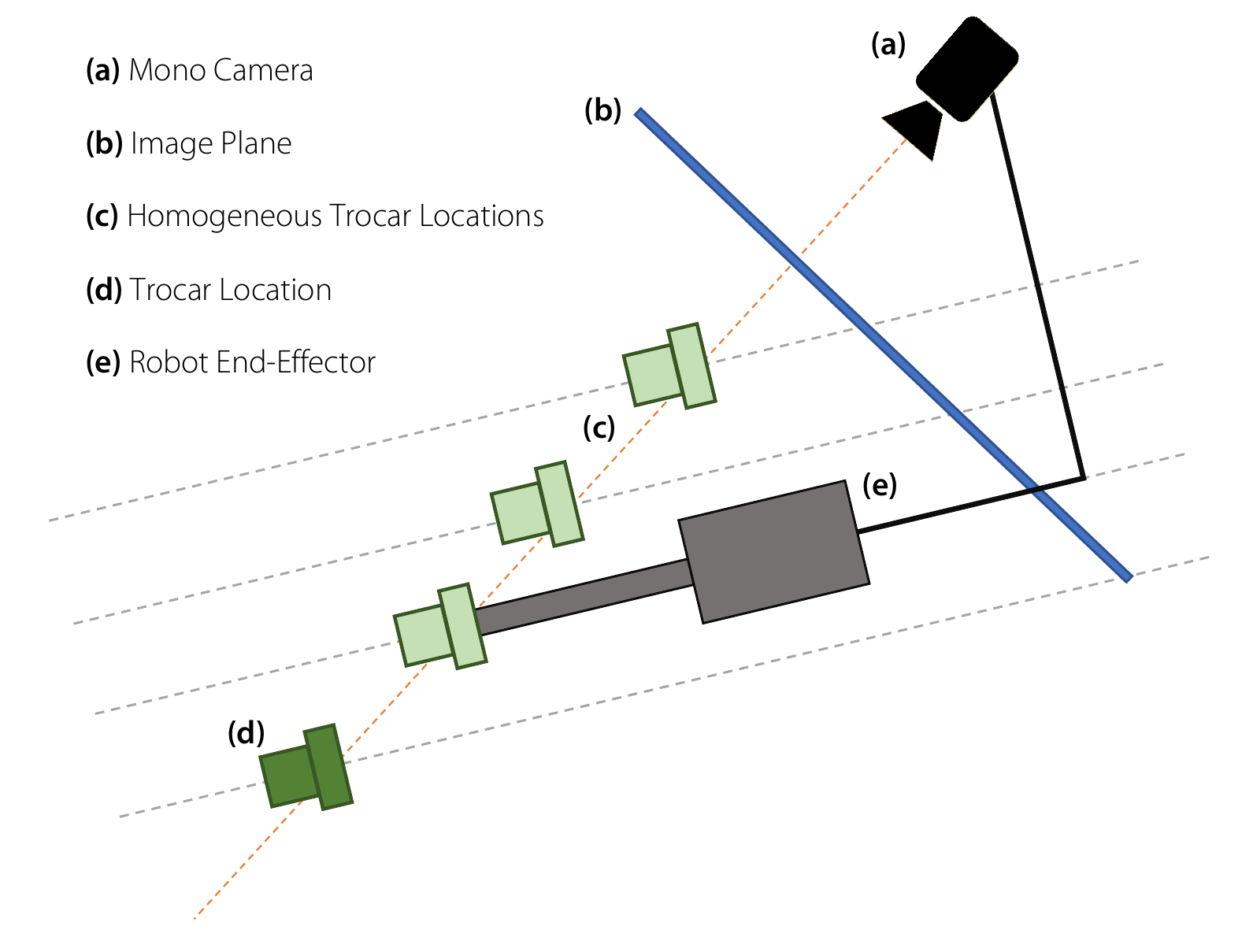}
    \caption{Trajectory of the tooltip, compared to the camera and trocar's homogeneous location, illustrated in a 2D projection.}
    \label{fig:RobotMove}
    \vspace{-0.4cm}
\end{figure}

Due to the non-unique $Z$ value obtained from the TEP detection, it is not possible to apply a direct docking process. This creates the need for designing a method to dynamically find the appropriate trajectory for the robot. In this work we have designed a procedure, inspired by hummingbird's docking approach~\cite{goller2014hummingbirds}. We first align the tooltip and trocar orientation, followed by the alignment of the $XY$ translation of the tooltip with TEP. As drawn in Fig.~\ref{fig:RobotMove}, keeping the tooltip on the line which connects the trocar's entry location to the camera, we start approaching the trocar with an adaptive speed. With this approach we ensure the tooltip is always aligned with the trocar, and the tooltip's projected position is aligned with the TEP's projected position in each image plane. By iteratively following this approach, we can also compensate for small movements of trocar.
\begin{figure}[H]
    \centering
    \includegraphics[width=0.8\linewidth]{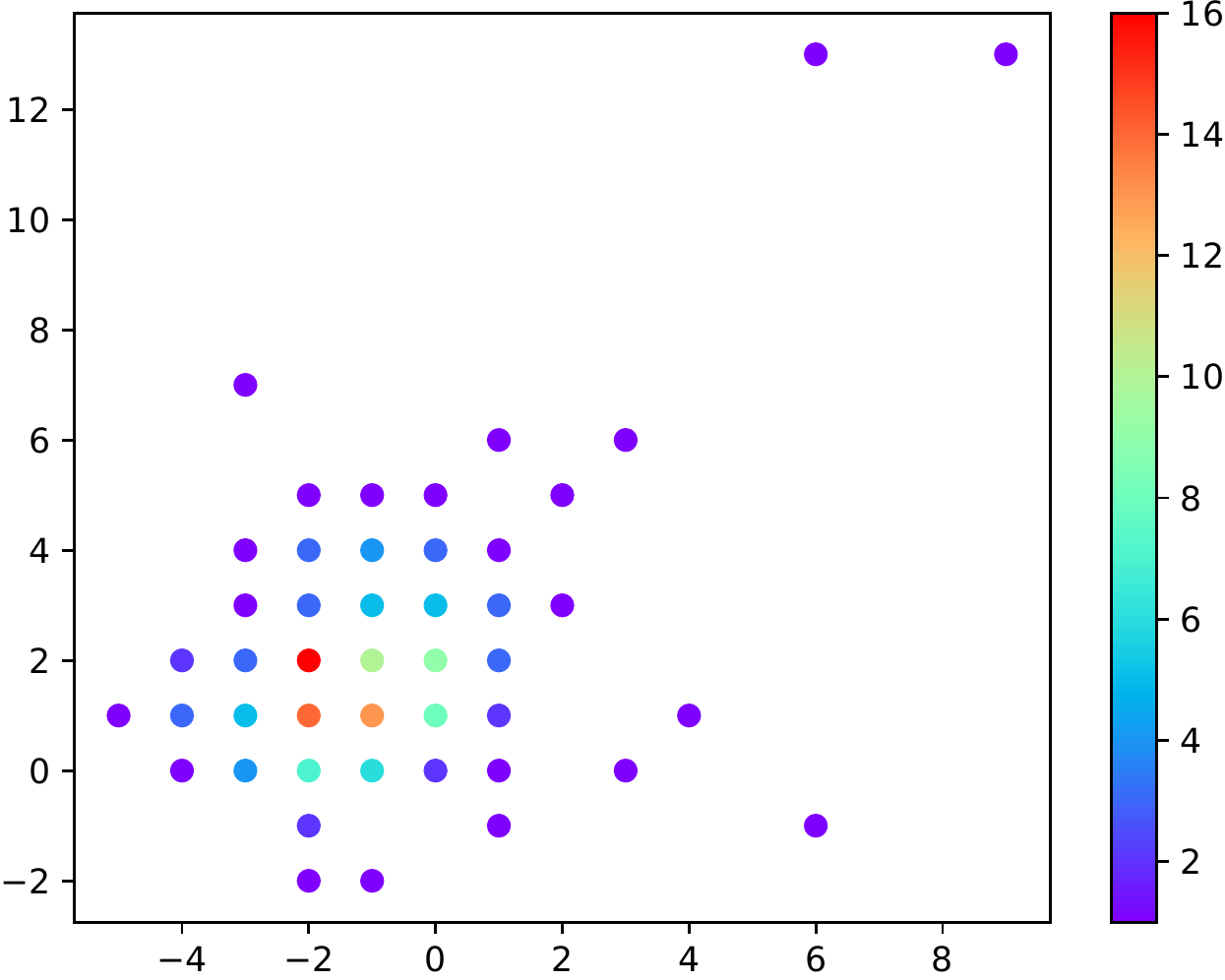}
    \caption{The offset between the ground truth TEP and the detected TEP in x and y direction in pixel. The color scheme indicates the number of times each offset was attained. To aid illustration, two outliers with euclidean distances to the ground truth location of 320 and 216 pixel were omitted.}
    \label{fig:DistancesXYPx}
    \vspace{-0.4cm}
\end{figure}
\section{Experiments and Results}
\label{sec:experiment}
To validate our proposed system, we first separately evaluate its sub-components and finally the overall autonomous trocar docking performance.
In the following, we demonstrate the accuracy of the TEP detection and trocar orientation estimation. 
We finally recreate a surgical scenario using phantom eyes and demonstrate the validity of our system by reporting the success rate of our method.
We additionally show that the autonomous docking procedure does not increase the operation time compared to the manual joystick-based approach.
Besides the quantitative results presented in the following sections, we provide qualitative visual results of the TEP and orientation estimation as well as the autonomous docking procedure as supplementary materials along with this paper.

\subsection{Trocar Entry Point Detection}
\begin{figure}[H]
    \centering
    \includegraphics[width=0.9\linewidth]{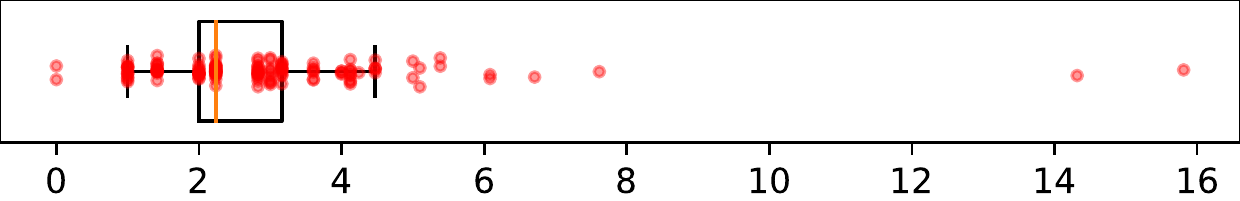}
    \caption{The euclidean distances between the estimated TEPs and the respective ground truth locations in pixel. Similar to Fig.~\ref{fig:DistancesXYPx}, two outliers were omitted for improved visualization.}
    \label{fig:DistancesEuclideanPX}
    \vspace{-0.4cm}
\end{figure}

To evaluate the detection of the TEP we use our test set $DS_{testTEP}$ consisting of 160 images, which were extracted from a video captured with our proposed setup.
The 2D image coordinates of the ground truth TEPs were manually annotated in the image frames.
The detected location of the TEP is then compared to its respective ground truth.
The overall achieved median, mean and standard deviation of the error in x and y image coordinates are 6.19, 2.82, and 30,29 pixel, respectively.
Given an image resolution of 1280$\times$720 pixels using our robot mounted camera, the offset error results in a mean of 0.74\%, a median of 0.28\%, and a standard deviation of 3.75\% from the annotated entry point. 
Fig.~\ref{fig:DistancesXYPx} illustrates the detection error along with the number of occurrences of each offset evaluated in our test set. 
The corresponding euclidean distances in pixels are visualized in Fig.~\ref{fig:DistancesEuclideanPX}.
The comparably high standard deviation is caused by two outliers, of which one showed a euclidean pixel distance of 320 and 216 pixels, respectively.

\begin{figure}[H]
\vspace{0.2cm}
\label{fig:poseresult}
    \centering
    \includegraphics[width=1\linewidth]{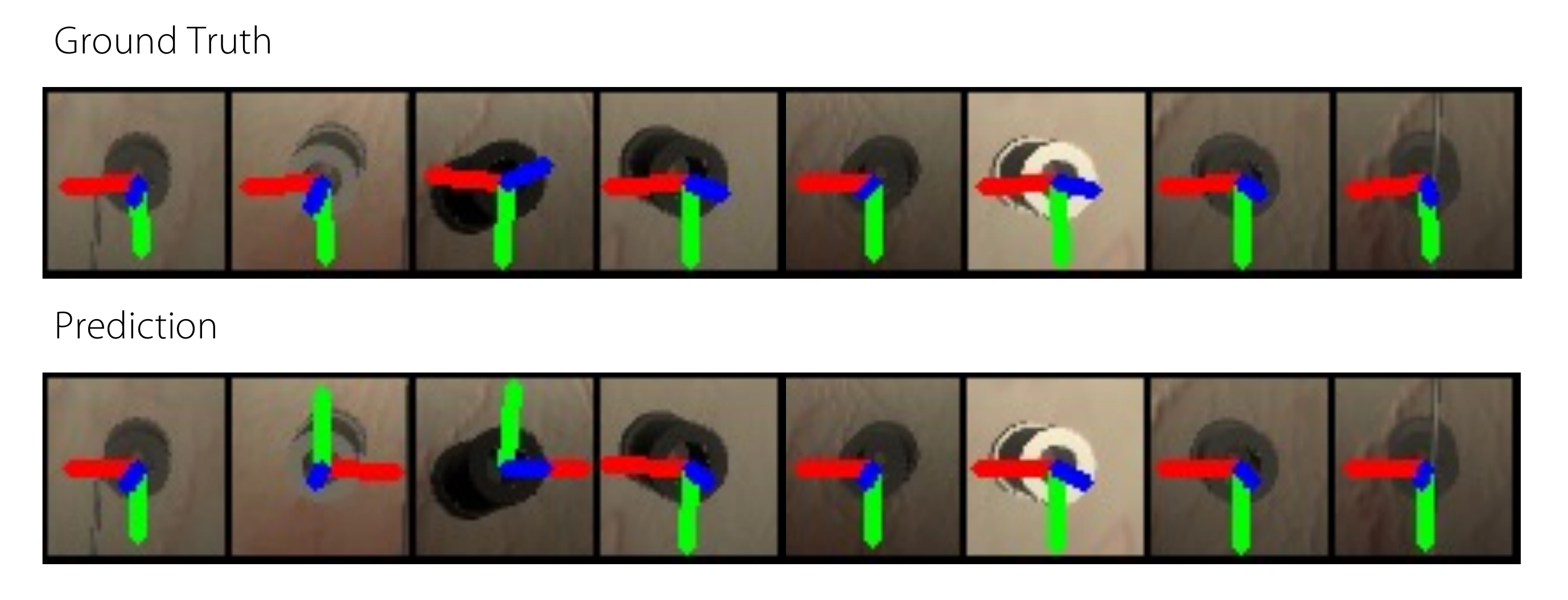}
    \caption{Comparison between the ground truths and the pose regression network on the test set of the synthetic dataset.}
\vspace{-0.4cm}
\end{figure}

\subsection{Trocar Orientation Estimation}
\begin{figure}[H]
    \label{fig:poseerror}
    \vspace{0.3cm}
    \centering
    \includegraphics[width=.9\linewidth]{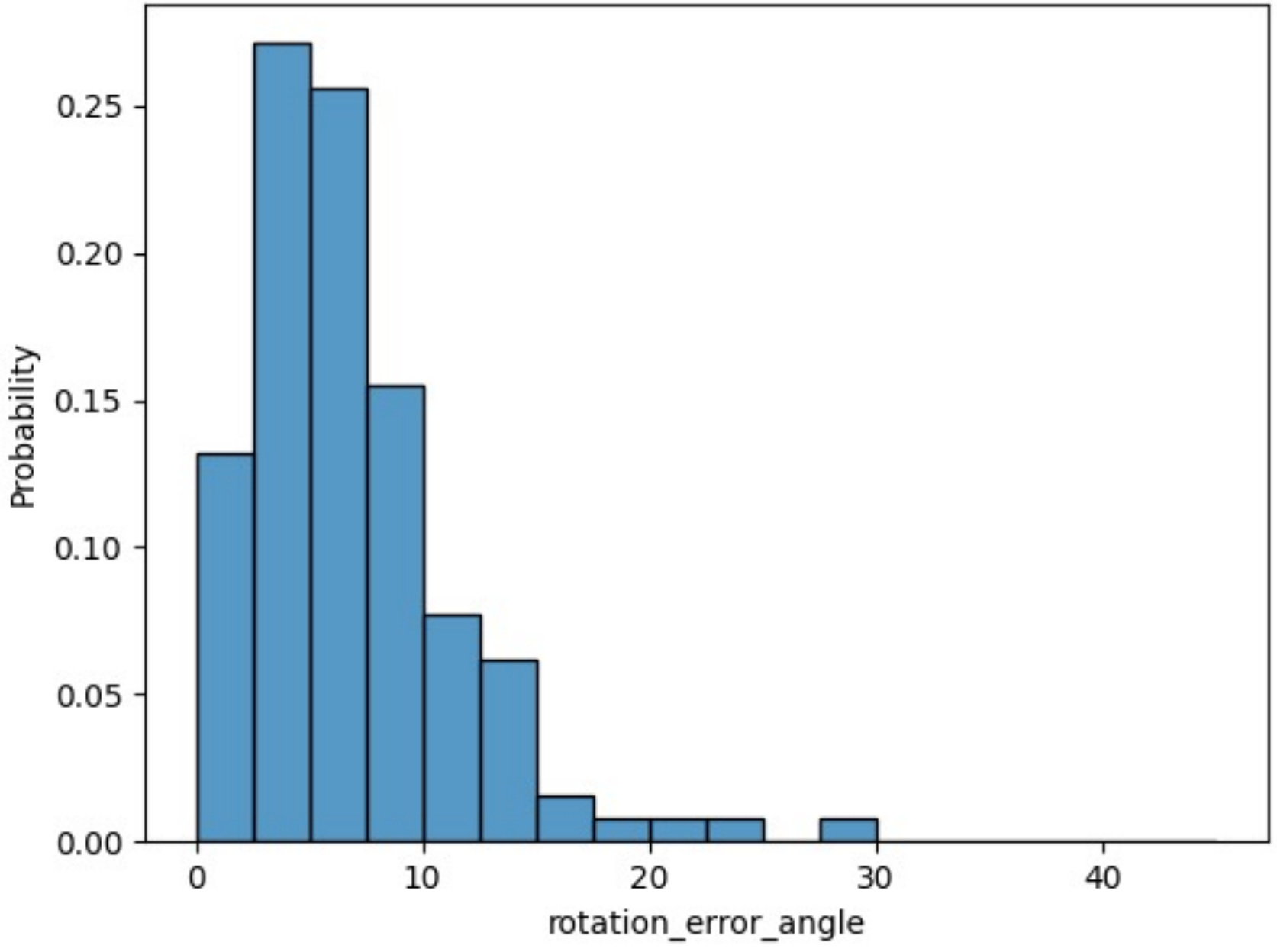}
    \caption{Pose estimation error on test set of the real dataset}
\end{figure}

After training the pose estimation model on the $DS_{synthetic}$ and fine-tuning on $DS_{realPoseTrain}$, we evaluated the results on the real images of our test set $DS_{realPoseTest}$. 
As illustrated in the Fig.~\ref{fig:poseerror}, the model has achieved an accuracy of 80\% to estimate the trocar orientation below $10^{\circ}$ error, and 94\% below $15^{\circ}$, which is achieved by fine tuning on a small set of real data. 
It is worth mentioning that the generation of a real dataset with ground truth orientation of the microsurgical trocar is challenging and prone to small errors due to the marker-based estimation.
Therefore, this quantitative evaluation is limited to the quality of the generated ground truth orientation.

\subsection{Autonomous Trocar Docking}
To validate our entire system for autonomous trocar docking, we simulate a surgical environment consisting of the robotic setup and a VR Eye from Philips Studio, which is attached to a head phantom.
For each trial we initially position the robot at a randomized location near the target trocar, ensuring that the docking target is located within the robot workspace and that the TEP can be reached during the docking process.
From the initial position, the robot then aligns the instrument with the trocar orientation and moves the tip towards the TEP as described in Section~\ref{sec:TrajectoryPlanning}.
Our experiments showed that out of 11 attempts, the robot was able to successfully reach the TEP and perform docking in 10 cases.
In one case, the instrument touched the edge of the trocar and missed the entry point.
However, this was due to a hand-eye calibration error between the instrument tip and the camera caused by a deformation of the tooltip.
The average time to completion 10 successful autonomous trials was 35.1 seconds with a standard deviation of 3.2 seconds.
For comparison, two biomedical engineers trained to control the robot with a joystick performed the docking task with the same setup. 
The manual alignment took on average 40.8 seconds but showed a higher variance in the time to completion with a standard deviation of 12.75 seconds.
The maximum time required for autonomous docking was 41 seconds compared to 57 seconds for manual alignment


\section{Discussion and future work}
\label{sec:discussion}
In our current experiments, one constraint of our method is the working space of the microsurgical robot.
To perform successful docking, the robot has to be positioned in proximity to the trocar, so that the instrument can be aligned with the trocar and the entry point can be reached.
However, the challenging act of aligning and inserting the instrument is still performed autonomously by the robot.
When using a robot with a larger working space, autonomous docking could also be performed from a greater starting distance.

In this work, we used a sub-retinal cannula (23G with 40G tip) with a straight tip, therefore, the introduction of the cannula into the trocar was performed by a final "Z" movement, following the proper docking alignment. For bent-tip cannulas, this method can be extended for more complex introduction trajectories. Here, if the trocar is out of detection range (due to movement of the eye during the procedure, or coverage by an external objects or blood), the docking procedure is stopped until the trocar is back to the detection range. 

As our experiments have shown, the current setup is able to perform autonomous trocar docking and insert the instrument into the eye. Once the instrument has reached the desired insertion depth, the process is manually stopped. In future works, we will investigate suitable stopping criteria to autonomously insert specific tools to the desired depth within the eye (e.g. when the tip appears in the microscopic view).
Additionally, our approach can be extended to dynamically adjust for changing trocar position during the insertion process.
We will also further analyze the force applied during autonomous robotic trocar docking and compare it to the force applied during conventional manual insertion, as autonomous trocar docking could lower the force applied on the sclera, thereby reducing patient trauma. 

\section{Conclusion}
\label{sec:conclusion}
In this paper we proposed an autonomous docking system inspired by the natural behaviour of hummingbirds for minimally invasive robotic surgery. Goller, Altshuler et al. in~\cite{goller2014hummingbirds} and~\cite{goller2017visual}, presented "Feeding hummingbirds use vision, not touch, to hover at flowers". They showed hummingbirds are identifying their target entry point and after approaching they fix their head and proceed in "Z" direction to insert their beak into a flower. 
Learning from this natural phenomena, our system for autonomous trocar docking and instrument insertion is based on a camera mounted to the endeffector of a micron-precision robot.
The approach first obtains the position of the TEP and subsequently estimates the trocar's orientation from RGB images.
The robot then aligns instrument with the trocar orientation and autonomously performs the docking procedure.
Our experiments have shown that the TEP can be detected with a clinical grade accuracy with a mean euclidean error of 3 pixel given an image resolution of 1280 $\times$ 720 pixel.
The detection of this point is extremely important for the subsequent pose estimation, since the input for the pose estimation network comprises a small region around the estimated TEP.
Further experiments were performed to validate our system by evaluating the entire autonomous docking procedure using a conventional 23G trocar mounted on a surgical training phantom eye:
Successful trocar docking could be achieved with high repeatability and an average time of 35 seconds, which indicates the potential of our method.
We consider this work as a proof that the proposed method for autonomous robotic trocar docking is a valid approach towards automatizing a task, which in robotic surgery would otherwise require additional cognitive effort from the surgeon.
\newpage
\bibliographystyle{IEEEtran}
{\scriptsize
\balance
\bibliography{MyCollection}
}
\end{document}